\documentclass{article} % For LaTeX2e
\usepackage[preprint]{colm2025_conference}

\usepackage{microtype}
\usepackage{hyperref}
\usepackage{url}
\usepackage{booktabs}
\usepackage{multirow}
\usepackage{float}
\usepackage{amsmath}
\usepackage{comment}
\usepackage{xspace}

\usepackage{lineno}

\usepackage{graphicx}

\definecolor{darkblue}{rgb}{0, 0, 0.5}
\hypersetup{colorlinks=true, citecolor=darkblue, linkcolor=darkblue, urlcolor=darkblue}

\title{An Analysis of Decoding Methods for LLM-based Agents for Faithful Multi-Hop Question Answering}

\author{Alexander Murphy\textsuperscript{1}\thanks{Correspondence to: \texttt{alexandermurphy784@gmail.com}} \quad Mohd Sanad Zaki Rizvi\textsuperscript{1} \quad Aden Haussmann\textsuperscript{2}\thanks{Work done while at the University of Edinburgh.} \quad Ping Nie\textsuperscript{2}\\
\textbf{Guifu Liu}\textsuperscript{1} \quad \textbf{Aryo Pradipta Gema}\textsuperscript{1} \quad \textbf{Pasquale Minervini}\textsuperscript{1,3} \\
\textsuperscript{1}University of Edinburgh \qquad
\textsuperscript{2}Independent \qquad
\textsuperscript{3}Miniml.AI
}

\def\ie{{\em i.e.,}\xspace}

\begin{document}

\ifcolmsubmission
\linenumbers
\fi

\maketitle

\begin{abstract}
Large Language Models (LLMs) frequently produce factually inaccurate outputs --- a phenomenon known as hallucination --- which limits their accuracy in knowledge-intensive NLP tasks.
Retrieval-augmented generation and agentic frameworks such as \textit{Reasoning and Acting} (ReAct) can address this issue by giving the model access to external knowledge. % integrating external knowledge into LLMs.
However, LLMs often fail to remain faithful to retrieved information.
Mitigating this is critical, especially if LLMs are required to reason about the retrieved information.
Recent research has explored training-free decoding strategies to improve the faithfulness of model generations.
We present a systematic analysis of how the combination of the ReAct framework and decoding strategies (\ie DeCoRe, DoLa, and CAD) can influence the faithfulness of LLM-generated answers.
Our results show that combining an agentic framework for knowledge retrieval with decoding methods that enhance faithfulness can increase accuracy on the downstream Multi-Hop Question Answering tasks.
For example, we observe an F1 increase from 19.5 to 32.6 on HotpotQA when using ReAct and DoLa.
%
%%% We don't really know, since these models are black-boxes
%
%This is due to the LLM relying more on retrieved context, and closer adherence to the prompt format required by the agent.
%
%%% The following sentence does not provide any information
%
%Based on our analysis, we provide insights into how and when this combination is most effective.
%
\end{abstract}

\section{Introduction}

\begin{figure}[t!]
  \centering
  \includegraphics[width=1.0\textwidth]{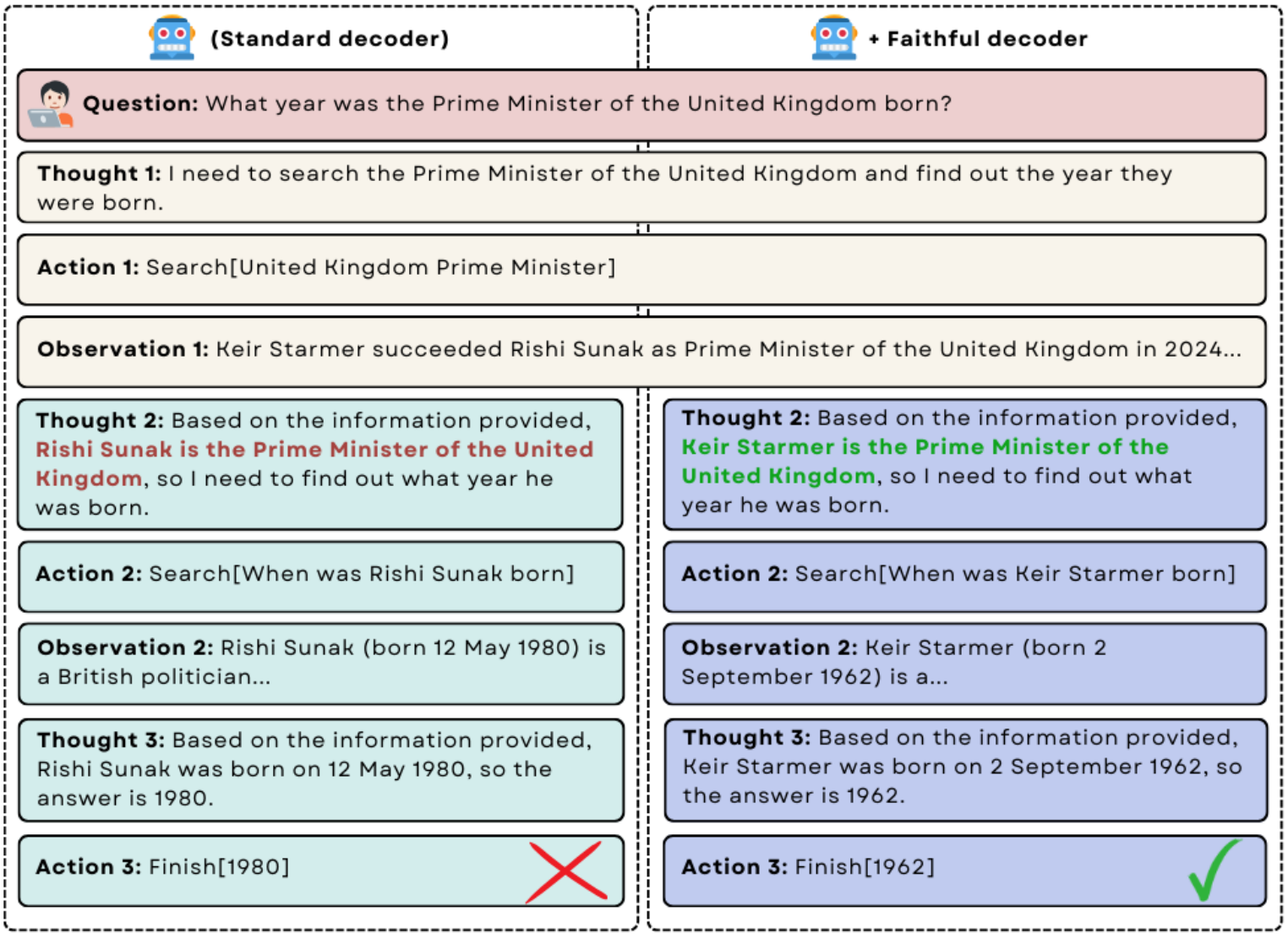}
  \caption{An example showing the difference between ReAct using standard vs. faithful decoding.
  The Thoughts, Actions and Observations depend on previous steps --- in the standard decoder, a mistake in Thought 2 propagates throughout the reasoning chain, yielding a wrong answer.
  The faithful decoder stays faithful to the observation, yielding the correct answer.
  %
  %An LLM could make a mistake seen in Thought 2 if, \eg it was trained before July 2024, when Rishi Sunak was the Prime Minister of the UK.
  } \label{fig:diagram_compare}
  \vspace{-20pt}
\end{figure}

Large Language Models (LLMs) are exceptional in generating fluent and coherent natural language, making them powerful tools for a wide range of applications~\citep{Radford2019LanguageMA, NEURIPS2020_1457c0d6, Wei2021FinetunedLM, 10.5555/3600270.3602281}.
However, an ongoing challenge with LLMs is their tendency to generate factually inaccurate outputs, a phenomenon known as \emph{hallucination}~\citep{10.1145/3571730, rawte2023surveyhallucinationlargefoundation, Zhang2023SirensSI, li-etal-2024-dawn, hong2024hallucinations}.
This tendency limits their reliability and accuracy, especially in knowledge-intensive NLP tasks such as Open-Domain Question Answering (ODQA).
Retrieval-Augmented Generation~\citep[RAG;][]{10.5555/3495724.3496517} can be used to condition the generation of an LLM on external knowledge and is widely used to improve the factual accuracy of the model in knowledge-intensive NLP tasks~\citep{li2024enhancingllmfactualaccuracy, Ayala_2024}.
In a basic RAG setup for question answering (QA), a retriever system uses the question to retrieve relevant texts from an external knowledge source, and the model generates an answer based on the retrieved information.
However, this ``one-step retrieve-and-read'' approach is insufficient for more complex QA tasks that require multi-hop reasoning~\citep{trivedi-etal-2023-interleaving}.
The Reasoning and Acting~\citep[ReAct;][]{yao2023react} framework addresses this by making the setup iterative and \emph{agentic}.
It enables the LLM to generate interpretable reasoning traces and task-specific actions in an interleaved manner.
Frameworks like ReAct can be used to provide the LLM with access to a search engine that can retrieve verifiable and up-to-date external knowledge.
However, even with access to external tools and knowledge, LLMs can still fail to remain faithful to such knowledge.
These events are referred to as \emph{faithfulness hallucinations}~\citep{10.1145/3703155} and may happen when the implicit knowledge in the parameters of the LLM conflicts with the provided context \citep{shi-etal-2024-trusting, zhao2024steering}.
%
% We argue that although ReAct and similar frameworks perform well on ODQA benchmarks and reduce hallucinations to some extent \citep{yao2023react, trivedi-etal-2023-interleaving}, they are not fundamental solutions to the problem of hallucination.
% %
% No matter how sophisticated the document retrieval or tool use framework is, the LLM itself will remain a performance bottleneck if it cannot be faithful to the retrieved information.
%
Mitigating this is critical, especially in applications where LLMs are required to reason from the provided context, such as multi-hop QA tasks.
Recent works explored training-free decoding methods that aim to improve the faithfulness of model generations \citep{shi-etal-2024-trusting, gema2024decoredecodingcontrastingretrieval, chuang2024doladecodingcontrastinglayers, Chen2024InContextSA}.
These decoding methods %use various contrastive or constrained strategies 
are characterised by how they influence the probability distribution of the next predicted token during the generation process.
We present the first systematic analysis of how the combination of an agentic framework (\ie ReAct) and different faithful decoding strategies influences the accuracy of LLM-generated answers to open-domain questions requiring multi-hop reasoning (see Figure \ref{fig:diagram_compare} for an example).
%
% We also evaluate one-step retrieval and direct prompting as baselines.
%
We specifically focus on evaluating ReAct without fine-tuning and selecting decoding methods that do not require training, as training-free approaches lower the barrier to entry for leveraging LLM agents.
% This deliberate choice is motivated by training-free approaches for downstream tasks.
%
% We chose to use training-free approaches to reduce the barrier of entry to leverage agentic frameworks.

Our findings show that the combination of an agentic framework for knowledge retrieval, and decoding methods that enhance contextual faithfulness, increases accuracy in the downstream task of multi-hop ODQA. We observe consistent improvements in answer and retrieval metrics across datasets and models, with Answer F1 improvements of up to 13.1 and Recall improvements of up to 4.5.

\section{Methodology}

\subsection{Frameworks}

% ReAct
ReAct \citep{yao2023react} is a reasoning framework that consists of 3 steps, where the system retrieves and reasons in an interleaved manner. The agent chooses an \emph{action} to perform, from which it receives an \emph{observation}. Then, the agent must \emph{reason} about the newly collected observation to decide on the next action. It repeats these steps until it decides that it has found an answer. Similarly to \citet{yao2023react}, we give the ReAct agent tools to \emph{Search} for a paragraph and to \emph{Lookup} the first sentence in an article containing a specific keyword. The article is always the one that the latest retrieved paragraph belongs to. The search tool returns the most relevant paragraph to the query specified by the agent.

As comparisons we use OneR, or \textit{One-step Retriever} and Direct prompting. OneR retrieves one paragraph of context, using the same retriever as ReAct, but with the question as the query. We limit it to one paragraph only, so as to simulate the same tool access as ReAct. Direct prompting doesn't get any external context at all and has to answer based just on the question.

\subsection{Hallucinations Reduction Decoding Strategies}

% CAD
\textbf{Context-Aware Decoding}~\citep[CAD;][]{shi-etal-2024-trusting} is a decoding method that computes token probabilities with and without a given context $c$.
%
%Token probabilities are then increased if the added context makes them a lot more probable than without, aiming to give more importance to the context the model is given than the knowledge encoded in the model weights.
%
When predicting the next token, CAD then increases the likelihood of the tokens selected by the model with $c$ while decreasing the likelihood of the tokens selected by the model without $c$, under the assumption that the former will tend to be more faithful to the context $c$ than the latter.
%
%We choose to incorporate CAD in the ReAct framework because it has been shown to be more faithful towards the provided context. This should assist ReAct in generating thoughts and actions which are more closely related to the retrieved observations.
%
Results in \citet{shi-etal-2024-trusting} show that generations produced with CAD tend to be more faithful to the context, which motivates us to integrate it within the ReAct framework.
%

% DoLa
\textbf{Decoding by Contrasting Layers}~\citep[DoLa;][]{chuang2024doladecodingcontrastinglayers} uses the insight that factual knowledge in LLMs is often localised to specific layers of the model.
It derives its next-token distributions by contrasting the logits obtained from early and later exiting layers, assuming that later layers will tend to generate more factually accurate predictions.
%
%This process aims to upweight probabilities generated by the more factual later layers.
%
We test its compatibility with ReAct since it has been shown to consistently improve Chain-of-Thought reasoning and instruction-following ability of LLMs~\citep{chuang2024doladecodingcontrastinglayers}.

% DeCoRe
\textbf{Decoding by Contrasting Retrieval Heads}~\citep[DeCoRe;][]{gema2024decoredecodingcontrastingretrieval} is a decoding method that dynamically contrasts outputs from a base LLM and an LLM with masked \emph{retrieval heads} --- attention heads that are responsible for extracting contextual information~\citep{wu2025retrieval}.
It uses conditional entropy of the model’s next-token distribution to guide the contrastive decoding mechanism. DeCoRe has been shown to improve open-book question-answering performance in a classic RAG setup.
Note that while CAD requires the context to be explicitly provided in the input, DoLa and DeCoRe leverage internal components of the transformer architecture (\ie layers and attention heads, respectively) in their decoding strategies that are not dependent on separately provided context.
\subsection{Metrics}
%

%\arh{I think it's weird we start with Answer EM when that isn't even in the main text. We should explain Answer F1 and Answer Support Recall before the appendix metrics. In fact I'd say those should only be explained in the appendix} 

\textbf{Answer F1} is a weighted average of precision and recall of the predicted answer compared to the gold answer at the token level, representing how accurate the final answer the agent generates is.
In our results in \autoref{App-Results}, we also include \emph{Answer EM}, which is only one if the gold answer matches the predicted answer exactly and zero otherwise.
\textbf{Answer Support Recall} is the fraction of questions for which the correct answer appears in one of the documents retrieved by ReAct.
We use this metric to evaluate how good the agent is at retrieving correct documents.
Note that in multi-hop QA, supporting documents do not necessarily contain the answer, as multi-hop QA may require multiple retrieval steps to find the answer.
For this reason, we include Paragraph F1 in \autoref{App-Results}, which compares all supporting paragraphs with the retrieved ones.

For both Answer F1 and Answer Support Recall, we use common normalization techniques, such as applying lowercasing and removing common articles and punctuation, before calculating the metric.

\section{Experimental Setup}

We now describe the experiment setup, including datasets and detailed descriptions of the ReAct components used.
All of the experiments are run three times on the same test sets, with seeds and technical details listed in \autoref{App-repro}.

\subsection{Datasets}

In our experiments we use the following three datasets. For each dataset we create a test set of 500 questions and a knowledge base to use for retrieval.

% HotpotQA
\textbf{HotpotQA}~\citep{yang2018hotpotqa} contains 113,000 Wikipedia-based, multi-hop question-answer pairs. This means that a question requires multiple documents to answer. For each question it contains the supporting paragraphs from each document needed to answer the question, which we use to evaluate retrieval metrics. As our knowledge base we use the provided 2017 Wikipedia corpus.

% 2WikiMultihopQA
\textbf{2WikiMultihopQA}~\citep{xanh2020_2wikimultihop} is a multi-hop question answering dataset, with questions, relevant context and answers collected from Wikipedia. It was specifically designed to avoid questions that can bypass multi-hop reasoning and always require at least one hop to be solved. We combine all supporting and non-supporting paragraphs from all questions, into one knowledge base for retrieval.

% MuSiQue
\textbf{MuSiQue} \citep{trivedi2021musique} consists of questions, relevant context and answers collected from Wikipedia.
Similarly to 2WikiMultihopQA, it provides multi-hop question-answer pairs that cannot be "cheated" with shortcuts and require actual multi-step reasoning. It also contains questions that are up to 4-hop and that are supposed to be unanswerable.
As with 2WikiMultihopQA, we combine all supporting and non-supporting paragraphs from all questions into one knowledge base.

\subsection{Details of ReAct components}

We run experiments with Qwen2-7b-Instruct \citep{yang2024qwen2technicalreport} and Llama3-8b-Instruct \citep{llama3} as the backbone LLM. We use BM25 \citep{bm25} implemented in Elasticsearch 7.10.2 as our retriever. We use few-shot prompts (seven examples) for ReAct, OneR and Direct to help outputs adhere to the correct format, similar to the setup \citet{yao2023react} use in their question answering experiments (exact prompts in \autoref{App-Prompts}). We use default hyperparameters for the decoders.

The ReAct agent has two available actions for interacting with the knowledge base for HotpotQA: \emph{Search} uses a given query to search the knowledge base, and \emph{Lookup} looks for the next paragraph containing a keyword in the current article.
2WikiMultihopQA and MuSiQue do not have the \emph{Lookup} action since the knowledge bases for these datasets do not consist of coherent articles. The prompt instructs the LLM on which tools are available and how to call them. We use chat templates to structure the reasoning and actions of the LLM and the observations it gets, noting that this improves performance considerably compared to no chat templates. If the LLM does not adhere to the \emph{Thought-Action-Observation} format specified in the prompt, the generation for the question is stopped, but we still use the previously generated output to evaluate retrieval metrics.

\section{Results}

In this section we discuss the results of the experiments run across frameworks, datasets, decoders and models. Our main evaluation metrics, Answer F1 and Answer Support Recall, are shown in Tables \ref{tab:f1} and \ref{tab:recall}, respectively. Additional metrics---Answer EM and Paragraph F1---are in \autoref{App-Results}.

%Our results show that, faithful decoding strategies combined with ReAct, largely improve performance on both Answer F1 and Answer Support Recall when compared to standard decoding, and conduct a qualitative analysis of the internal mechanisms that cause this. Additionally, we compare performance differences between the two LLMs and propose hypotheses for these. Finally, we show that faithful decoding does not improve Answer F1 for OneR, as OneR fails to retrieve the paragraph which contains the answer in one step, rendering faithful decoding futile. These results validate faithful decoding strategies as a method to improve faithfulness in Agentic, multi-step QA, and provide insight into when they add the most value, and when they are not useful.

\begin{table}[H]
    \centering
    \begin{tabular}{llccc}
    \toprule
    \textbf{Framework}           & \textbf{Decoding}                & \textbf{HotpotQA} & \textbf{2WikiMultihopQA} & \textbf{MuSiQue} \\
    \midrule
    \multicolumn{5}{c}{\textit{Qwen2-7b-Instruct}} \\
    \midrule
    \multirow{1}{*}{Direct}      & Standard                         & 20.3 $\pm$ 0.0 & 28.4 $\pm$ 0.0        & 8.9 $\pm$ 0.0 \\ \midrule
    \multirow{4}{*}{OneR}        & Standard                         & \textbf{27.0 $\pm$ 0.0} & \textbf{21.0 $\pm$ 0.0}        & \textbf{9.5 $\pm$ 0.0}\\
                                 & CAD                              & 21.7 $\pm$ 0.0 & 12.2 $\pm$ 0.0        & 8.5 $\pm$ 0.0\\
                                 & DoLa                             & 20.2 $\pm$ 0.0 & 12.6 $\pm$ 0.0        & 9.1 $\pm$ 0.0\\
                                 & $\text{DeCoRe}_{\text{entropy}}$ & 23.3 $\pm$ 0.0 & 17.6 $\pm$ 0.0        & 9.2 $\pm$ 0.0\\ \midrule
    \multirow{4}{*}{ReAct}       & Standard                         & 19.5 $\pm$ 0.0 & 43.1 $\pm$ 0.1        & 10.3 $\pm$ 0.0\\
                                 & CAD                              & 28.5 $\pm$ 0.2 & \textbf{51.8 $\pm$ 0.1}        & 11.3 $\pm$ 0.0\\
                                 & DoLa                             & \textbf{32.6 $\pm$ 0.5} & 42.4 $\pm$ 0.2        & 11.9 $\pm$ 0.1\\
                                 & $\text{DeCoRe}_{\text{entropy}}$ & 26.7 $\pm$ 0.2 & 49.3 $\pm$ 0.4        & \textbf{12.0 $\pm$ 0.1}\\
    \midrule
    \multicolumn{5}{c}{\textit{Llama3-8b-Instruct}} \\
    \midrule
    \multirow{1}{*}{Direct}      & Standard                         & 28.9 $\pm$ 0.0 & 33.5 $\pm$ 0.0        & 12.3 $\pm$ 0.0\\ \midrule
    \multirow{4}{*}{OneR}        & Standard                         & 39.5 $\pm$ 0.0 & 35.7 $\pm$ 0.0        & 15.0 $\pm$ 0.0\\
                                 & CAD                              & 37.8 $\pm$ 0.0 & 29.7 $\pm$ 0.0        & 13.4 $\pm$ 0.0\\
                                 & DoLa                             & 39.9 $\pm$ 0.0 & \textbf{36.3 $\pm$ 0.0}        & 14.9 $\pm$ 0.0\\
                                 & $\text{DeCoRe}_{\text{entropy}}$ & \textbf{40.0 $\pm$ 0.0} & 35.8 $\pm$ 0.0        & \textbf{15.1 $\pm$ 0.0}\\ \midrule
    \multirow{4}{*}{ReAct}       & Standard                         & 34.7 $\pm$ 0.0 & 54.6 $\pm$ 0.1        & 11.7 $\pm$ 0.0\\
                                 & CAD                              & 27.0 $\pm$ 0.0 & 51.4 $\pm$ 0.2        & 11.2 $\pm$ 0.1\\
                                 & DoLa                             & \textbf{35.7 $\pm$ 0.0} & \textbf{55.1 $\pm$ 0.1}      & 11.5 $\pm$ 0.3\\
                                 & $\text{DeCoRe}_{\text{entropy}}$ & 29.8 $\pm$ 0.0 & 46.1 $\pm$ 0.1        & \textbf{12.2 $\pm$ 0.7}\\
    \bottomrule
    \end{tabular}
    \caption{Answer F1 grouped by model, framework, decoder and dataset. The highest F1 per group is in bold.}
    \label{tab:f1}
\end{table}

\begin{table}[H]
    \centering
    \begin{tabular}{llccc}
    \toprule
    \textbf{Framework}           & \textbf{Decoding}                & \textbf{HotpotQA} & \textbf{2WikiMultihopQA} & \textbf{MuSiQue} \\
    \midrule
    \multicolumn{5}{c}{\textit{Qwen2-7b-Instruct}} \\
    \midrule
    \multirow{1}{*}{OneR}        & All                              & 27.8 $\pm$ 0.0 & 19.0 $\pm$ 0.0        & 8.2 $\pm$ 0.0\\ \midrule
    \multirow{4}{*}{ReAct}       & Standard                         & 44.3 $\pm$ 0.5 & 66.4 $\pm$ 0.0        & 11.0 $\pm$ 0.3\\
                                 & CAD                              & 43.0 $\pm$ 0.0 & 66.8 $\pm$ 0.0        & \textbf{11.4 $\pm$ 0.7}\\
                                 & DoLa                             & \textbf{46.6 $\pm$ 0.7} & 68.4 $\pm$ 0.0        & 10.8 $\pm$ 0.7\\
                                 & $\text{DeCoRe}_{\text{entropy}}$ & 41.9 $\pm$ 0.1 & \textbf{70.9 $\pm$ 0.1}        & 10.3 $\pm$ 0.1\\
    \midrule
    \multicolumn{5}{c}{\textit{Llama3-8b-Instruct}} \\
    \midrule
    \multirow{1}{*}{OneR}        & All                              & 27.8 $\pm$ 0.0 & 19.0 $\pm$ 0.0        & 8.2 $\pm$ 0.0\\ \midrule
    \multirow{4}{*}{ReAct}       & Standard                         & \textbf{49.0 $\pm$ 0.0} & 83.8 $\pm$ 0.0        & 11.9 $\pm$ 0.1\\
                                 & CAD                              & 43.4 $\pm$ 0.0 & 80.5 $\pm$ 0.1        & 12.1 $\pm$ 0.1\\
                                 & DoLa                             & \textbf{49.0 $\pm$ 0.0} & 83.6 $\pm$ 0.0        & 12.9 $\pm$ 0.1\\
                                 & $\text{DeCoRe}_{\text{entropy}}$ & 45.6 $\pm$ 0.0 & \textbf{85.0 $\pm$ 0.0}        & \textbf{13.2 $\pm$ 0.0}\\\bottomrule
    \end{tabular}
    \caption{Answer Support Recall grouped by model, framework, decoder and dataset. The highest Answer Support Recall per group is in bold. 'All' under decoding indicates that all decoders get the same value, since they do not influence retrieval for OneR.}
    \label{tab:recall}
\end{table}

\subsection{Faithful Decoding Improves Answering, Retrieval \& Format Adherence}

\paragraph{Answer F1 and Answer Support Recall Improvements}In 6 out of 6 cases, ReAct combined with a faithful decoder increases Answer F1, showing that improvements hold across datasets and models. The most substantial increase is 13.1, from 19.5 to 32.6, for Qwen with DoLa on HotpotQA (a relative increase of over 50\%). For Llama we see smaller increases of up to 1.3. However, on Llama especially, we also see some decoders that reduce Answer F1, sometimes significantly. For Qwen on the other hand there only is one case of a faithful decoder reducing performance, by just under 1 percentage point.

Answer Support Recall improves in 5 out of 6 cases, with a tie for Llama on HotpotQA. The greatest improvement is for Qwen with DeCoRe on 2WikiMultihopQA, increasing by 4.5. For Llama there is a smaller improvement, of up to 1.3. Similarly to Answer F1, there also are decoders that reduce Recall, especially for Llama, though there always is at least one decoder that causes an improvement or tie.

\paragraph{Prompt Format Adherence Improvements}

% \subsection{Faithful Decoding Improves Prompt Format Adherence in ReAct} \label{adhere}

To measure the impact faithful decoding has on ReAct following the \textit{Thought-Action-Observation} format specified in the prompt (see \autoref{App-Prompts} for exact prompts), we measure the proportion of reasoning traces which contain the \textit{Finish[...]} keyword in the final output.

\begin{figure}[t]
    \centering
    \includegraphics[width=1\linewidth]{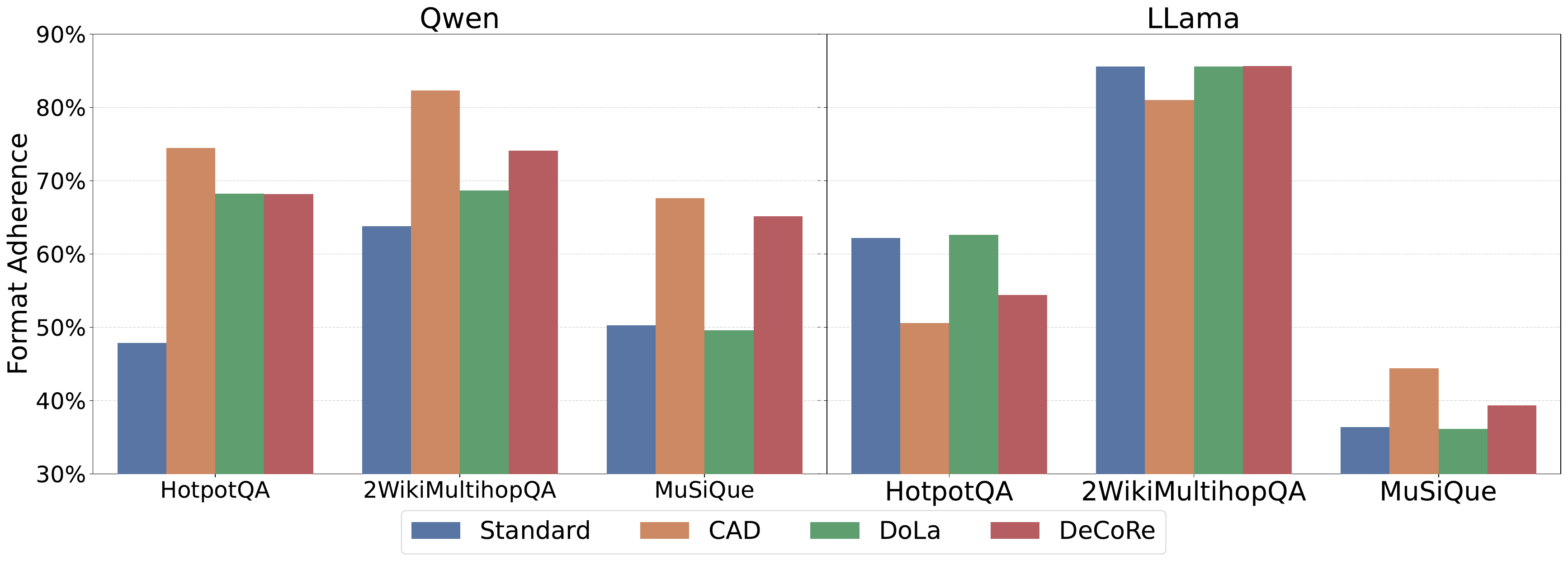}
    \caption{The proportion of questions for which a complete reasoning trace was created with ReAct (measured by how many have the \textit{Finish[...]} keyword in their final output).}
    \label{fig:completion}
\end{figure}

Figure \ref{fig:completion} shows that in most cases, format adherence is either improved or matched when using faithful decoding. For Qwen, across all datasets, CAD increases the number of questions adhering to the format the most. The greatest increase is seen in HotpotQA, raising the proportion of complete reasoning traces from 47.9\% to 74.5\%. This correlates with the significant increase in Answer F1 seen on HotpotQA with all faithful decoders applied to Qwen, where Answer F1 improved from 19.5 with Standard decoding, to 28.5 with CAD and 32.6 with DoLa.

Adherence improves on Llama as well, but to a lesser extent and less consistently across decoders. Although there is no large difference between the standard decoder and the best faithful decoders on HotpotQA and 2WikiMultihopQA (with some decoders even decreasing adherence), there is a significant increase in adherence on MuSiQue, with CAD raising the proportion of complete reasoning traces from 36.4\% to 44.4\%. 

Interestingly, even though CAD improves format adherence the most for Qwen, this does not directly correlate to the highest Answer F1. This suggests that not all decoders improve performance in the same way, and some are better at improving format adherence such as CAD, while others improve inherent factuality (such as DoLa focusing on factual transformer layers).

These improvements in adherence could reduce the need for fine-tuning ReAct, which \citet{yao2023react} use to ensure format adherence. We show the raw values for adherence in \autoref{App-Results}.

% \subsection{Qualitative Analysis}

\paragraph{Qualitative Analysis}

Our qualitative analysis of output reasoning and action traces supports the quantitative results. Firstly, we notice a reduced amount of factual mistakes. Secondly, faithful decoders help generate outputs in the correct ReAct format more consistently.

\begin{figure}[t]
  \centering
  \includegraphics[width=1.0\textwidth]{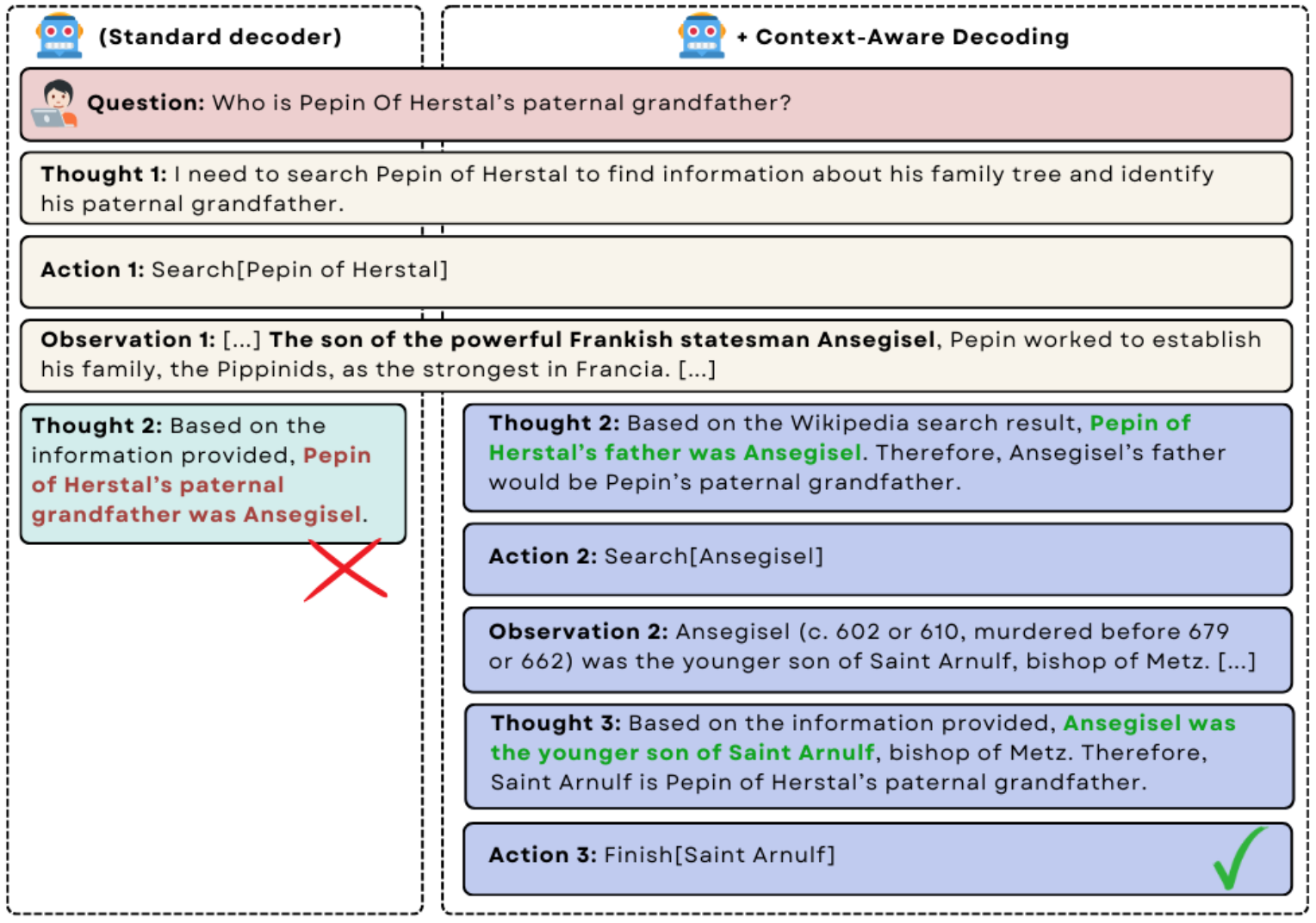}
  \caption{An example from 2WikiMultihopQA where faithful decoding improves upon standard decoding with Qwen2-7b-Instruct. Standard decoding is not faithful to the retrieved context and generates an incorrect thought, derailing the reasoning chain. CAD helps the agent stay faithful to context and leads it to the correct solution.}
  \label{fig:diagram_example}
\end{figure}

In Figure \ref{fig:diagram_example}, we compare the output trace of the Standard decoder with that of CAD. We observe that the reasoning chain of CAD values the context more highly, correctly identifying the passage \textit{"Ansegisel's father would be Pepin's paternal grandfather"} in its second Thought, compared to the Standard decoder which thinks that \textit{"Pepin of Herstal's paternal grandfather was Ansegisel"}, a classic faithfulness hallucination. This small distinction leads the agent with CAD to take another action and \textit{Search} for the father of \textit{"Ansegisel"}, while the Standard agent does not, as it thinks it has found the correct answer immediately at the first step.

In addition to being more faithful to the context, the CAD agent also formats the actions correctly, while the Standard agent leaves its answer in the Thought, neglecting the final \textit{Finish} action.

\subsection{Influence of LLMs' Prior Knowledge}

We observe that ReAct, using Llama, does not surpass OneR's Answer F1 on HotpotQA or MuSiQue, despite a significantly higher Answer Support Recall than OneR. 
% This shows that under certain circumstances, the benefits of ReAct may be limited. \citet{yao2023react} show that finetuning the model on correct reasoning chains can improve ReAct performance considerably compared to OneR, as one major reason for ReAct's low performance is incorrectly formatted reasoning chains.
Llama also receives a high Answer F1 with Direct prompting, using no context at all.
These seemingly contradictory results point to the possibility that Llama is leveraging pre-existing knowledge stored implicitly in its parameters. 

Interestingly, Qwen does not do as well without context, receiving low Answer F1 scores with OneR and Direct prompting, compared to Llama. Yet this disparity is less pronounced when using ReAct. This suggests that the agentic framework may level the playing field between different LLMs to some extent. Especially with faithful decoding methods, an LLM's dependence on having seen the answer in it's training data is reduced, as more weight is put on the provided context. Using faithful decoding on a finetuned model which conforms to the ReAct format more consistently and uses context better, is potentially a training-free way of increasing performance even more.

\subsection{Limitations of Faithful Decoding in the Absence of Relevant Context}

We also notice that for OneR, across both models, the faithful decoders do not yield substantial or consistent improvements in Answer F1 over Standard decoding.
In fact, for Qwen, we observe performance degradation across all datasets.
This observation is attributed to OneR's limitations in retrieving the paragraph containing the correct answer, illustrated by OneR's low Answer Support Recall in \autoref{tab:recall}. The inadequacy of single-step retrieve-and-read systems for complex, multi-hop ODQA benchmarks has been documented in previous research \citep{trivedi-etal-2023-interleaving}. In comaprison to OneR, ReAct has a significantly higher Answer Support Recall in all experiments, and thus, Answer F1 is also increased.

These findings underscore a crucial principle: if the LLM is not provided with, or is unable to retrieve, a context that contains the correct answer, then even perfect adherence to the retrieved context will not lead to improved answer accuracy and may even degrade performance.
The benefit of faithful decoding in this setup is, therefore, contingent on the availability of relevant information.

\subsection{No Decisive Superiority Among Faithful Decoders}

In 3 out of 6 experiments, DoLa achieves the largest improvement for ReAct, when looking at Answer F1. For format adherence, CAD improves the most by a significant amount in 3 out of 6 cases. Nevertheless, our analysis does not reveal a single faithful decoder that consistently outperforms the others or any definitive patterns as to when one decoder performs better than another. While at least one faithful decoder demonstrated improvements over, or parity with, Standard decoding across all datasets with ReAct, there was no discernable pattern of any decoder performing particularly well or poorly on a certain dataset.

\section{Related works}

\paragraph{Other retrieval \& reasoning frameworks}

Interleaving Retrieval with Chain-of-Thought Reasoning~\citep[IRCoT;][]{trivedi-etal-2023-interleaving} enhances the conventional RAG process by interleaving the retrieval of information with the generation of reasoning steps (Chain-of-Thoughts, or CoT). At each step of the CoT, retrieval is guided by reasoning, and the retrieved context is integrated to improve reasoning in turn. This interleaved approach is similar to ReAct, however IRCoT is only used for multi-hop QA, whereas ReAct can be used for other knowledge-intensive reasoning tasks, such as fact verification \citep{yao2023react}. Additionally, IRCoT can only retrieve and read at each step, but ReAct chooses task-specific actions to take. IRCoT significantly outperformed one-step retrieve-and-read on ODQA benchmarks \citep{trivedi-etal-2023-interleaving}.

\paragraph{Other decoding strategies}

Autocontrastive Decoding~\citep[ACD;][]{gera-etal-2023-benefits}, contrasts higher and lower layers of the model to improve the quality of generations, similarly to DoLa. However, \citet{gera-etal-2023-benefits} noted that ACD seemed to increase the tendency of LLMs to hallucinate. This makes it unsuitable for our study, and is therefore excluded. Activation Decoding \citep{Chen2024InContextSA} uses entropy to quantify the 'sharpness' of in-context hidden states, and was shown to mitigate hallucinations.

\paragraph{Detecting \& mitigating hallucination in RAG systems}

Regressing Decoupled External context score and Parametric knowledge score~\citep[ReDeEP;][]{Sun2024ReDeEPDH}, detects hallucination in RAG systems by 'disentangling' how models use their own parametric knowledge from how they use external sources. ReDeEP increases the accuracy of RAG hallucination detection. \citet{Sun2024ReDeEPDH} also introduce Add Attention Reduce FFN (AARF). AARF improves truthfulness by modulating the contributions of \textit{Knowledge FFNs (Feed Forward Networks)} and \textit{Copying Heads} in the LLM, stopping the model from relying too heavily on the former and underutilizing the latter, which they claim is the cause of hallucinations. Layer-wise Relevance Propagation for RAG~\citep[LRP4RAG;][]{hu2024lrp4ragdetectinghallucinationsretrievalaugmented} uses the Layer-wise Relevance Propagation (LRP) algorithm to detect hallucinations. Both ReDeEP and LRP4RAG outperformed previous baselines at hallucination detection on the RAGTruth corpus \citep{niu-etal-2024-ragtruth}. Our study distinguishes itself by being the first systematic analysis of several decoding methods for improving faithfulness and their interplay with RAG and agentic knowledge-intensive reasoning systems.

\section{Conclusion}

This study addresses the challenge of mitigating hallucination in LLMs when answering open-domain questions requiring multi-hop reasoning with an agentic RAG setup. We investigate the impact of various decoding strategies on the faithfulness of LLM-generated answers within the ReAct agentic framework and compare this to one-step retrieval and direct prompting baselines. Our analysis reveals that faithful decoding methods, such as DeCoRe, DoLa and CAD, enhance question answering performance when integrated with ReAct. This is at least partly due to the LLM being more faithful to retrieved context and adhering more closely to the prompt format required by the agent. This improvement underscores the importance of contextual faithfulness in achieving accurate and reliable LLM outputs, as the models are better able to utilise retrieved paragraphs when generating answers.

% \arh{These are too long for inclusion in the conclusion, but I think they summarise the other main findings nicely. Maybe we can add them as a summary at the end of results/discussion?}

% However, our findings also demonstrate that the efficacy of faithful decoding is contingent upon the broader system architecture. In contrast to ReAct, the one-step retrieval approach (OneR) did not exhibit performance gains from faithful decoding. This highlights the critical importance of employing a system that is appropriate for the complexity of the task. As observed in our experiments, applying faithful decoding to a system with inherent limitations, such as OneR's poor performance on multi-hop question answering, will not improve results.

% Furthermore, while faithful decoding strategies generally improve performance, our analysis did not identify a single universally superior decoding method. The optimal choice of decoding strategy appears to be task-dependent, necessitating empirical evaluation to determine the most effective option.

% Finally, faithful decoding proves most beneficial when the system is accessing information external to the LLM's training data, e.g. proprietary knowledge or internal documentation. In such scenarios, it is crucial to select an appropriate system architecture for the task at hand and to engage in empirical testing to identify the most suitable decoding strategy.

% Doesn't count towards page count
\section{Ethics Statement}

The use of the various decoding strategies presented in combination with ReAct, aims to reduce hallucinations and improve the faithfulness and correctness of agents. This has the potential to significantly increase agents' trustworthiness.

However, harmful contexts could become more pronounced in the agent output, and malicious actors could use this fact to manipulate agents.
Risks should be managed accordingly, and precautions should be taken on a case-by-case basis.

Our code is publicly available in order to allow users to gain insight in the use of decoding strategies applied to agents and help them make the safest and best decision for their use-case.

%\section*{Author Contributions}
%If you'd like to, you may include a section for author contributions, as is done in many journals. This is optional and at the discretion of the authors.

\section*{Acknowledgments}
%Use unnumbered first-level headings for the acknowledgements. All acknowledgements, including those to funding agencies, go at the end of the paper.

Aryo Pradipta Gema was supported by the United Kingdom Research and Innovation (grant EP/S02431X/1), UKRI Centre for Doctoral Training in Biomedical AI at the University of Edinburgh, School of Informatics.
Pasquale Minervini was partially funded by ELIAI, an industry grant from Cisco, and a donation from Accenture LLP.
This work was supported by the Edinburgh International Data Facility (EIDF), the Data-Driven Innovation Programme at the University of Edinburgh, and Netmind.AI.

\bibliography{colm2025_conference}

\begin{thebibliography}{31}
\providecommand{\natexlab}[1]{#1}
\providecommand{\url}[1]{\texttt{#1}}
\expandafter\ifx\csname urlstyle\endcsname\relax
  \providecommand{\doi}[1]{doi: #1}\else
  \providecommand{\doi}{doi: \begingroup \urlstyle{rm}\Url}\fi

\bibitem[Ayala \& Bechard(2024)Ayala and Bechard]{Ayala_2024}
Orlando Ayala and Patrice Bechard.
\newblock Reducing hallucination in structured outputs via retrieval-augmented generation.
\newblock In \emph{Proceedings of the 2024 Conference of the North American Chapter of the Association for Computational Linguistics: Human Language Technologies (Volume 6: Industry Track)}, pp.\  228–238. Association for Computational Linguistics, 2024.
\newblock \doi{10.18653/v1/2024.naacl-industry.19}.
\newblock URL \url{http://dx.doi.org/10.18653/v1/2024.naacl-industry.19}.

\bibitem[Brown et~al.(2020)Brown, Mann, Ryder, Subbiah, Kaplan, Dhariwal, Neelakantan, Shyam, Sastry, Askell, Agarwal, Herbert-Voss, Krueger, Henighan, Child, Ramesh, Ziegler, Wu, Winter, Hesse, Chen, Sigler, Litwin, Gray, Chess, Clark, Berner, McCandlish, Radford, Sutskever, and Amodei]{NEURIPS2020_1457c0d6}
Tom Brown, Benjamin Mann, Nick Ryder, Melanie Subbiah, Jared~D Kaplan, Prafulla Dhariwal, Arvind Neelakantan, Pranav Shyam, Girish Sastry, Amanda Askell, Sandhini Agarwal, Ariel Herbert-Voss, Gretchen Krueger, Tom Henighan, Rewon Child, Aditya Ramesh, Daniel Ziegler, Jeffrey Wu, Clemens Winter, Chris Hesse, Mark Chen, Eric Sigler, Mateusz Litwin, Scott Gray, Benjamin Chess, Jack Clark, Christopher Berner, Sam McCandlish, Alec Radford, Ilya Sutskever, and Dario Amodei.
\newblock Language models are few-shot learners.
\newblock In H.~Larochelle, M.~Ranzato, R.~Hadsell, M.F. Balcan, and H.~Lin (eds.), \emph{Advances in Neural Information Processing Systems}, volume~33, pp.\  1877--1901. Curran Associates, Inc., 2020.
\newblock URL \url{https://proceedings.neurips.cc/paper_files/paper/2020/file/1457c0d6bfcb4967418bfb8ac142f64a-Paper.pdf}.

\bibitem[Chen et~al.(2024)Chen, Xiong, Liu, Wu, Xiao, Gao, and He]{Chen2024InContextSA}
Shiqi Chen, Miao Xiong, Junteng Liu, Zhengxuan Wu, Teng Xiao, Siyang Gao, and Junxian He.
\newblock In-context sharpness as alerts: An inner representation perspective for hallucination mitigation, 2024.
\newblock URL \url{https://arxiv.org/abs/2403.01548}.

\bibitem[Chuang et~al.(2024)Chuang, Xie, Luo, Kim, Glass, and He]{chuang2024doladecodingcontrastinglayers}
Yung-Sung Chuang, Yujia Xie, Hongyin Luo, Yoon Kim, James~R. Glass, and Pengcheng He.
\newblock Dola: Decoding by contrasting layers improves factuality in large language models.
\newblock In \emph{The Twelfth International Conference on Learning Representations}, 2024.
\newblock URL \url{https://openreview.net/forum?id=Th6NyL07na}.

\bibitem[Gema et~al.(2024)Gema, Jin, Abdulaal, Diethe, Teare, Alex, Minervini, and Saseendran]{gema2024decoredecodingcontrastingretrieval}
Aryo~Pradipta Gema, Chen Jin, Ahmed Abdulaal, Tom Diethe, Philip Teare, Beatrice Alex, Pasquale Minervini, and Amrutha Saseendran.
\newblock Decore: Decoding by contrasting retrieval heads to mitigate hallucinations, 2024.
\newblock URL \url{https://arxiv.org/abs/2410.18860}.

\bibitem[Gera et~al.(2023)Gera, Friedman, Arviv, Gunasekara, Sznajder, Slonim, and Shnarch]{gera-etal-2023-benefits}
Ariel Gera, Roni Friedman, Ofir Arviv, Chulaka Gunasekara, Benjamin Sznajder, Noam Slonim, and Eyal Shnarch.
\newblock The benefits of bad advice: Autocontrastive decoding across model layers.
\newblock In Anna Rogers, Jordan Boyd-Graber, and Naoaki Okazaki (eds.), \emph{Proceedings of the 61st Annual Meeting of the Association for Computational Linguistics (Volume 1: Long Papers)}, pp.\  10406--10420, Toronto, Canada, July 2023. Association for Computational Linguistics.
\newblock \doi{10.18653/v1/2023.acl-long.580}.
\newblock URL \url{https://aclanthology.org/2023.acl-long.580/}.

\bibitem[Grattafiori et~al.(2024)Grattafiori, Abhimanyu~Dubey, Pandey, Kadiana, Al-Dahle, Letman, Mathur, Schelten, Vaughan, Yang, and et~al.]{llama3}
Aaron Grattafiori, Abhinav~Jauhri Abhimanyu~Dubey, Abhinav Pandey, Abhishek Kadiana, Ahmad Al-Dahle, Aiesha Letman, Akhil Mathur, Alan Schelten, Alex Vaughan, Amy Yang, and Angela~Fan et~al.
\newblock The llama 3 herd of models, 2024.
\newblock URL \url{https://arxiv.org/abs/2407.21783}.

\bibitem[Ho et~al.(2020)Ho, Duong~Nguyen, Sugawara, and Aizawa]{xanh2020_2wikimultihop}
Xanh Ho, Anh-Khoa Duong~Nguyen, Saku Sugawara, and Akiko Aizawa.
\newblock Constructing a multi-hop {QA} dataset for comprehensive evaluation of reasoning steps.
\newblock In \emph{Proceedings of the 28th International Conference on Computational Linguistics}, pp.\  6609--6625, Barcelona, Spain (Online), December 2020. International Committee on Computational Linguistics.
\newblock URL \url{https://www.aclweb.org/anthology/2020.coling-main.580}.

\bibitem[Hong et~al.(2024)Hong, Gema, Saxena, Du, Nie, Zhao, Perez-Beltrachini, Ryabinin, He, Fourrier, et~al.]{hong2024hallucinations}
Giwon Hong, Aryo~Pradipta Gema, Rohit Saxena, Xiaotang Du, Ping Nie, Yu~Zhao, Laura Perez-Beltrachini, Max Ryabinin, Xuanli He, Cl{\'e}mentine Fourrier, et~al.
\newblock The hallucinations leaderboard--an open effort to measure hallucinations in large language models.
\newblock \emph{arXiv preprint arXiv:2404.05904}, 2024.

\bibitem[Hu et~al.(2024)Hu, Sun, and Zhang]{hu2024lrp4ragdetectinghallucinationsretrievalaugmented}
Haichuan Hu, Yuhan Sun, and Quanjun Zhang.
\newblock Lrp4rag: Detecting hallucinations in retrieval-augmented generation via layer-wise relevance propagation, 2024.
\newblock URL \url{https://arxiv.org/abs/2408.15533}.

\bibitem[Huang et~al.(2025)Huang, Yu, Ma, Zhong, Feng, Wang, Chen, Peng, Feng, Qin, and Liu]{10.1145/3703155}
Lei Huang, Weijiang Yu, Weitao Ma, Weihong Zhong, Zhangyin Feng, Haotian Wang, Qianglong Chen, Weihua Peng, Xiaocheng Feng, Bing Qin, and Ting Liu.
\newblock A survey on hallucination in large language models: Principles, taxonomy, challenges, and open questions.
\newblock \emph{ACM Trans. Inf. Syst.}, 43\penalty0 (2), January 2025.
\newblock ISSN 1046-8188.
\newblock \doi{10.1145/3703155}.
\newblock URL \url{https://doi.org/10.1145/3703155}.

\bibitem[Ji et~al.(2023)Ji, Lee, Frieske, Yu, Su, Xu, Ishii, Bang, Madotto, and Fung]{10.1145/3571730}
Ziwei Ji, Nayeon Lee, Rita Frieske, Tiezheng Yu, Dan Su, Yan Xu, Etsuko Ishii, Ye~Jin Bang, Andrea Madotto, and Pascale Fung.
\newblock Survey of hallucination in natural language generation.
\newblock \emph{ACM Comput. Surv.}, 55\penalty0 (12), March 2023.
\newblock ISSN 0360-0300.
\newblock \doi{10.1145/3571730}.
\newblock URL \url{https://doi.org/10.1145/3571730}.

\bibitem[Lewis et~al.(2020)Lewis, Perez, Piktus, Petroni, Karpukhin, Goyal, K\"{u}ttler, Lewis, Yih, Rockt\"{a}schel, Riedel, and Kiela]{10.5555/3495724.3496517}
Patrick Lewis, Ethan Perez, Aleksandra Piktus, Fabio Petroni, Vladimir Karpukhin, Naman Goyal, Heinrich K\"{u}ttler, Mike Lewis, Wen-tau Yih, Tim Rockt\"{a}schel, Sebastian Riedel, and Douwe Kiela.
\newblock Retrieval-augmented generation for knowledge-intensive nlp tasks.
\newblock In \emph{Proceedings of the 34th International Conference on Neural Information Processing Systems}, NIPS '20, Red Hook, NY, USA, 2020. Curran Associates Inc.
\newblock ISBN 9781713829546.

\bibitem[Li et~al.(2024{\natexlab{a}})Li, Yuan, and Zhang]{li2024enhancingllmfactualaccuracy}
Jiarui Li, Ye~Yuan, and Zehua Zhang.
\newblock Enhancing llm factual accuracy with rag to counter hallucinations: A case study on domain-specific queries in private knowledge-bases, 2024{\natexlab{a}}.
\newblock URL \url{https://arxiv.org/abs/2403.10446}.

\bibitem[Li et~al.(2024{\natexlab{b}})Li, Chen, Ren, Cheng, Zhao, Nie, and Wen]{li-etal-2024-dawn}
Junyi Li, Jie Chen, Ruiyang Ren, Xiaoxue Cheng, Xin Zhao, Jian-Yun Nie, and Ji-Rong Wen.
\newblock The dawn after the dark: An empirical study on factuality hallucination in large language models.
\newblock In Lun-Wei Ku, Andre Martins, and Vivek Srikumar (eds.), \emph{Proceedings of the 62nd Annual Meeting of the Association for Computational Linguistics (Volume 1: Long Papers)}, pp.\  10879--10899, Bangkok, Thailand, August 2024{\natexlab{b}}. Association for Computational Linguistics.
\newblock \doi{10.18653/v1/2024.acl-long.586}.
\newblock URL \url{https://aclanthology.org/2024.acl-long.586/}.

\bibitem[Niu et~al.(2024)Niu, Wu, Zhu, Xu, Shum, Zhong, Song, and Zhang]{niu-etal-2024-ragtruth}
Cheng Niu, Yuanhao Wu, Juno Zhu, Siliang Xu, KaShun Shum, Randy Zhong, Juntong Song, and Tong Zhang.
\newblock {RAGT}ruth: A hallucination corpus for developing trustworthy retrieval-augmented language models.
\newblock In Lun-Wei Ku, Andre Martins, and Vivek Srikumar (eds.), \emph{Proceedings of the 62nd Annual Meeting of the Association for Computational Linguistics (Volume 1: Long Papers)}, pp.\  10862--10878, Bangkok, Thailand, August 2024. Association for Computational Linguistics.
\newblock \doi{10.18653/v1/2024.acl-long.585}.
\newblock URL \url{https://aclanthology.org/2024.acl-long.585/}.

\bibitem[Ouyang et~al.(2022)Ouyang, Wu, Jiang, Almeida, Wainwright, Mishkin, Zhang, Agarwal, Slama, Ray, Schulman, Hilton, Kelton, Miller, Simens, Askell, Welinder, Christiano, Leike, and Lowe]{10.5555/3600270.3602281}
Long Ouyang, Jeff Wu, Xu~Jiang, Diogo Almeida, Carroll~L. Wainwright, Pamela Mishkin, Chong Zhang, Sandhini Agarwal, Katarina Slama, Alex Ray, John Schulman, Jacob Hilton, Fraser Kelton, Luke Miller, Maddie Simens, Amanda Askell, Peter Welinder, Paul Christiano, Jan Leike, and Ryan Lowe.
\newblock Training language models to follow instructions with human feedback.
\newblock In \emph{Proceedings of the 36th International Conference on Neural Information Processing Systems}, NIPS '22, Red Hook, NY, USA, 2022. Curran Associates Inc.
\newblock ISBN 9781713871088.

\bibitem[Radford et~al.(2019)Radford, Wu, Child, Luan, Amodei, and Sutskever]{Radford2019LanguageMA}
Alec Radford, Jeff Wu, Rewon Child, David Luan, Dario Amodei, and Ilya Sutskever.
\newblock Language models are unsupervised multitask learners.
\newblock 2019.

\bibitem[Rawte et~al.(2023)Rawte, Sheth, and Das]{rawte2023surveyhallucinationlargefoundation}
Vipula Rawte, Amit Sheth, and Amitava Das.
\newblock A survey of hallucination in large foundation models, 2023.
\newblock URL \url{https://arxiv.org/abs/2309.05922}.

\bibitem[Robertson \& Zaragoza(2009)Robertson and Zaragoza]{bm25}
Stephen Robertson and Hugo Zaragoza.
\newblock The probabilistic relevance framework: Bm25 and beyond.
\newblock \emph{Found. Trends Inf. Retr.}, 3\penalty0 (4):\penalty0 333–389, April 2009.
\newblock ISSN 1554-0669.
\newblock \doi{10.1561/1500000019}.
\newblock URL \url{https://doi.org/10.1561/1500000019}.

\bibitem[Shi et~al.(2024)Shi, Han, Lewis, Tsvetkov, Zettlemoyer, and Yih]{shi-etal-2024-trusting}
Weijia Shi, Xiaochuang Han, Mike Lewis, Yulia Tsvetkov, Luke Zettlemoyer, and Wen-tau Yih.
\newblock Trusting your evidence: Hallucinate less with context-aware decoding.
\newblock In Kevin Duh, Helena Gomez, and Steven Bethard (eds.), \emph{Proceedings of the 2024 Conference of the North American Chapter of the Association for Computational Linguistics: Human Language Technologies (Volume 2: Short Papers)}, pp.\  783--791, Mexico City, Mexico, June 2024. Association for Computational Linguistics.
\newblock \doi{10.18653/v1/2024.naacl-short.69}.
\newblock URL \url{https://aclanthology.org/2024.naacl-short.69/}.

\bibitem[Sun et~al.(2025)Sun, Zang, Zheng, Xu, Zhang, Yu, Song, and Li]{Sun2024ReDeEPDH}
ZhongXiang Sun, Xiaoxue Zang, Kai Zheng, Jun Xu, Xiao Zhang, Weijie Yu, Yang Song, and Han Li.
\newblock Rede{EP}: Detecting hallucination in retrieval-augmented generation via mechanistic interpretability.
\newblock In \emph{The Thirteenth International Conference on Learning Representations}, 2025.
\newblock URL \url{https://openreview.net/forum?id=ztzZDzgfrh}.

\bibitem[Trivedi et~al.(2022)Trivedi, Balasubramanian, Khot, and Sabharwal]{trivedi2021musique}
Harsh Trivedi, Niranjan Balasubramanian, Tushar Khot, and Ashish Sabharwal.
\newblock {M}u{S}i{Q}ue: Multihop questions via single-hop question composition.
\newblock \emph{Transactions of the Association for Computational Linguistics}, 2022.

\bibitem[Trivedi et~al.(2023)Trivedi, Balasubramanian, Khot, and Sabharwal]{trivedi-etal-2023-interleaving}
Harsh Trivedi, Niranjan Balasubramanian, Tushar Khot, and Ashish Sabharwal.
\newblock Interleaving retrieval with chain-of-thought reasoning for knowledge-intensive multi-step questions.
\newblock In Anna Rogers, Jordan Boyd-Graber, and Naoaki Okazaki (eds.), \emph{Proceedings of the 61st Annual Meeting of the Association for Computational Linguistics (Volume 1: Long Papers)}, pp.\  10014--10037, Toronto, Canada, July 2023. Association for Computational Linguistics.
\newblock \doi{10.18653/v1/2023.acl-long.557}.
\newblock URL \url{https://aclanthology.org/2023.acl-long.557/}.

\bibitem[Wei et~al.(2022)Wei, Bosma, Zhao, Guu, Yu, Lester, Du, Dai, and Le]{Wei2021FinetunedLM}
Jason Wei, Maarten Bosma, Vincent Zhao, Kelvin Guu, Adams~Wei Yu, Brian Lester, Nan Du, Andrew~M. Dai, and Quoc~V Le.
\newblock Finetuned language models are zero-shot learners.
\newblock In \emph{International Conference on Learning Representations}, 2022.
\newblock URL \url{https://openreview.net/forum?id=gEZrGCozdqR}.

\bibitem[Wu et~al.(2025)Wu, Wang, Xiao, Peng, and Fu]{wu2025retrieval}
Wenhao Wu, Yizhong Wang, Guangxuan Xiao, Hao Peng, and Yao Fu.
\newblock Retrieval head mechanistically explains long-context factuality.
\newblock In \emph{The Thirteenth International Conference on Learning Representations}, 2025.
\newblock URL \url{https://openreview.net/forum?id=EytBpUGB1Z}.

\bibitem[Yang et~al.(2024)Yang, Yang, Hui, Zheng, Yu, Zhou, Li, Li, Liu, Huang, and et~al.]{yang2024qwen2technicalreport}
An~Yang, Baosong Yang, Binyuan Hui, Bo~Zheng, Bowen Yu, Chang Zhou, Chengpeng Li, Chengyuan Li, Dayiheng Liu, Fei Huang, and Guanting~Dong et~al.
\newblock Qwen2 technical report, 2024.
\newblock URL \url{https://arxiv.org/abs/2407.10671}.

\bibitem[Yang et~al.(2018)Yang, Qi, Zhang, Bengio, Cohen, Salakhutdinov, and Manning]{yang2018hotpotqa}
Zhilin Yang, Peng Qi, Saizheng Zhang, Yoshua Bengio, William~W. Cohen, Ruslan Salakhutdinov, and Christopher~D. Manning.
\newblock {HotpotQA}: A dataset for diverse, explainable multi-hop question answering.
\newblock In \emph{Conference on Empirical Methods in Natural Language Processing ({EMNLP})}, 2018.

\bibitem[Yao et~al.(2023)Yao, Zhao, Yu, Du, Shafran, Narasimhan, and Cao]{yao2023react}
Shunyu Yao, Jeffrey Zhao, Dian Yu, Nan Du, Izhak Shafran, Karthik~R Narasimhan, and Yuan Cao.
\newblock React: Synergizing reasoning and acting in language models.
\newblock In \emph{The Eleventh International Conference on Learning Representations}, 2023.
\newblock URL \url{https://openreview.net/forum?id=WE_vluYUL-X}.

\bibitem[Zhang et~al.(2023)Zhang, Li, Cui, Cai, Liu, Fu, Huang, Zhao, Zhang, Chen, Wang, Luu, Bi, Shi, and Shi]{Zhang2023SirensSI}
Yue Zhang, Yafu Li, Leyang Cui, Deng Cai, Lemao Liu, Tingchen Fu, Xinting Huang, Enbo Zhao, Yu~Zhang, Yulong Chen, Longyue Wang, Anh~Tuan Luu, Wei Bi, Freda Shi, and Shuming Shi.
\newblock Siren's song in the ai ocean: A survey on hallucination in large language models, 2023.
\newblock URL \url{https://arxiv.org/abs/2309.01219}.

\bibitem[Zhao et~al.(2024)Zhao, Devoto, Hong, Du, Gema, Wang, He, Wong, and Minervini]{zhao2024steering}
Yu~Zhao, Alessio Devoto, Giwon Hong, Xiaotang Du, Aryo~Pradipta Gema, Hongru Wang, Xuanli He, Kam-Fai Wong, and Pasquale Minervini.
\newblock Steering knowledge selection behaviours in llms via sae-based representation engineering.
\newblock \emph{arXiv preprint arXiv:2410.15999}, 2024.

\end{thebibliography}
\bibliographystyle{colm2025_conference}

\appendix

\section{Reproducibility}
\label{App-repro}

We run all our experiments three times with 1234, 3782 and 9539 as our seeds. We ran experiments on A100 and H100 GPUs, with ReAct experiments taking about 2 hours each and OneR and Direct experiments taking between 5 - 10 minutes each. Our code is publicly available\footnote{\href{https://github.com/awesomealex1/faithful-reasoning}{https://github.com/awesomealex1/faithful-reasoning}}. In addition to that we provide scripts to reproduce our results with correct seeds and results files used to create figures and tables. All models we used are publicly available from the Huggingface Transformers library and datasets are publicly available online. 

\section{Additional Results}
\label{App-Results}

In Tables \ref{tab:answer_em} and \ref{tab:para_f1} we record Answer EM and Paragraph F1 for our experiments. The results show that a faithful decoder with ReAct or exceeds Standard decoding for Answer EM on all datasets, and OneR does not consistently benefit from faithful decoders in Answer EM (Table \ref{tab:answer_em}). These results are consistent with the Answer F1 scores (Table \ref{tab:f1} in the main text) and reinforce the effective combination of ReAct with faithful decoders.

Unlike Answer Support Recall (Table \ref{tab:recall} in the main text), faithful decoders do not consistently beat standard decoding with ReAct in Paragraph F1 (Table \ref{tab:para_f1}). 

\begin{table}[H]
    \centering
    \begin{tabular}{llccc}
    \toprule
    \textbf{Framework}           & \textbf{Decoding}                & \textbf{HotpotQA} & \textbf{2WikiMultihopQA} & \textbf{MuSiQue} \\ \midrule
    \multicolumn{5}{c}{\textit{Qwen2-7b-Instruct}} \\
    \midrule
    \multirow{1}{*}{Direct}      & Standard                         & 10.8 $\pm$ 0.0 & 23.4 $\pm$ 0.0        & 2.2 $\pm$ 0.0\\ \midrule
    \multirow{4}{*}{OneR}        & Standard                         & \textbf{15.6 $\pm$ 0.0} & \textbf{21.0 $\pm$ 0.0}        & 2.6 $\pm$ 0.0\\
                                 & CAD                              & 11.0 $\pm$ 0.0 & 12.2 $\pm$ 0.0        & 2.0 $\pm$ 0.0\\
                                 & DoLa                             &  9.6 $\pm$ 0.0 & 12.6 $\pm$ 0.0        & \textbf{2.8 $\pm$ 0.0}\\
                                 & $\text{DeCoRe}_{\text{entropy}}$ & 12.6 $\pm$ 0.0 & 17.6 $\pm$ 0.0        & 2.4 $\pm$ 0.0\\ \midrule
    \multirow{4}{*}{ReAct}       & Standard                         & 14.3 $\pm$ 0.1 & 36.8 $\pm$ 0.0        & 5.9 $\pm$ 0.2\\
                                 & CAD                              & 21.9 $\pm$ 0.1 & 41.2 $\pm$ 0.2        & 5.6 $\pm$ 0.0\\
                                 & DoLa                             & \textbf{23.9 $\pm$ 0.2} & 35.6 $\pm$ 0.2        & 6.1 $\pm$ 0.1\\
                                 & $\text{DeCoRe}_{\text{entropy}}$ & 18.8 $\pm$ 0.3 & \textbf{42.5 $\pm$ 0.3}        & \textbf{6.5 $\pm$ 0.1}\\
    \midrule
    \multicolumn{5}{c}{\textit{Llama3-8b-Instruct}} \\
    \midrule
    \multirow{1}{*}{Direct}      & Standard                         & 21.4 $\pm$ 0.0 & 28.8 $\pm$ 0.0        & 4.2 $\pm$ 0.0\\ \midrule
    \multirow{4}{*}{OneR}        & Standard                         & 29.6 $\pm$ 0.0 & 30.6 $\pm$ 0.0        & 7.2 $\pm$ 0.0\\
                                 & CAD                              & 27.6 $\pm$ 0.0 & 24.0 $\pm$ 0.0        & 6.0 $\pm$ 0.0\\
                                 & DoLa                             & \textbf{30.2 $\pm$ 0.0} & \textbf{31.0 $\pm$ 0.0}        & 7.4 $\pm$ 0.0\\
                                 & $\text{DeCoRe}_{\text{entropy}}$ & 30.0 $\pm$ 0.0 & 30.4 $\pm$ 0.0        & \textbf{7.8 $\pm$ 0.0}\\ \midrule
    \multirow{4}{*}{ReAct}       & Standard                         & 26.8 $\pm$ 0.0 & \textbf{46.3 $\pm$ 0.1}        & \textbf{8.7 $\pm$ 0.1}\\
                                 & CAD                              & 20.8 $\pm$ 0.0 & 44.3 $\pm$ 0.2        & 7.0 $\pm$ 0.0\\
                                 & DoLa                             & \textbf{27.6 $\pm$ 0.0} & \textbf{46.3 $\pm$ 0.1}        & 8.2 $\pm$ 0.3\\
                                 & $\text{DeCoRe}_{\text{entropy}}$ & 22.8 $\pm$ 0.0 & 37.3 $\pm$ 0.1        & \textbf{8.7 $\pm$ 0.5}\\ \bottomrule
    \end{tabular}
    \caption{Answer EM grouped by model, framework, decoder and dataset. The highest EM per group is in bold.}
    \label{tab:answer_em}
\end{table}

\begin{table}[H]
    \centering
    \begin{tabular}{llccc}
    \toprule
    \textbf{Framework}           & \textbf{Decoding}                & \textbf{HotpotQA} & \textbf{2WikiMultihopQA} & \textbf{MuSiQue} \\ \midrule
    \multicolumn{5}{c}{\textit{Qwen2-7b-Instruct}} \\
    \midrule
    \multirow{1}{*}{OneR}        & All                              & 40.1 $\pm$ 0.0 & 40.5 $\pm$ 0.0        & 26.8 $\pm$ 0.0\\ \midrule
    \multirow{4}{*}{ReAct}       & Standard                         & 49.9 $\pm$ 0.1 & 83.4 $\pm$ 0.0        & \textbf{31.5 $\pm$ 0.3}\\
                                 & CAD                              & 49.4 $\pm$ 0.2 & 83.6 $\pm$ 0.1        & 30.8 $\pm$ 0.4\\
                                 & DoLa                             & \textbf{50.9 $\pm$ 0.5} & \textbf{84.7 $\pm$ 0.0}        & 31.3 $\pm$ 0.5\\
                                 & $\text{DeCoRe}_{\text{entropy}}$ & 49.1 $\pm$ 0.0 & 82.2 $\pm$ 0.1        & 31.1 $\pm$ 0.1\\
    \midrule
    \multicolumn{5}{c}{\textit{Llama3-8b-Instruct}} \\
    \midrule
    \multirow{1}{*}{OneR}        & All                              & 40.1 $\pm$ 0.0 & 40.5 $\pm$ 0.0        & 26.8 $\pm$ 0.0\\ \midrule
    \multirow{4}{*}{ReAct}       & Standard                         & \textbf{45.8 $\pm$ 0.0} & \textbf{89.3 $\pm$ 0.0}        & 27.2 $\pm$ 0.2\\
                                 & CAD                              & 44.1 $\pm$ 0.0 & 86.3 $\pm$ 0.1        & 27.4 $\pm$ 0.3\\
                                 & DoLa                             & 45.6 $\pm$ 0.0 & 88.3 $\pm$ 0.0        & 27.2 $\pm$ 0.1\\
                                 & $\text{DeCoRe}_{\text{entropy}}$ & 43.3 $\pm$ 0.0 & 86.8 $\pm$ 0.1        & \textbf{27.8 $\pm$ 0.1}\\ \bottomrule
    \end{tabular}
    \caption{Paragraph F1 grouped by model, framework, decoder and dataset. The highest F1 per group is in bold.}
    \label{tab:para_f1}
\end{table}

\begin{table}[H]
    \centering
    \begin{tabular}{llccc}
    \toprule
    \textbf{Framework}           & \textbf{Decoding}                & \textbf{HotpotQA} & \textbf{2WikiMultihopQA} & \textbf{MuSiQue} \\ \midrule
    \multicolumn{5}{c}{\textit{Qwen2-7b-Instruct}} \\
    \midrule
    \multirow{4}{*}{ReAct}       & Standard                         & 47.9 $\pm$ 0.5 & 63.8 $\pm$ 0.2        & 50.3 $\pm$ 0.5\\
                                 & CAD                              & \textbf{74.5 $\pm$ 0.2} & \textbf{82.3 $\pm$ 0.1}        & \textbf{67.6 $\pm$ 0.3}\\
                                 & DoLa                             & 68.2 $\pm$ 0.1 & 68.7 $\pm$ 0.1        & 49.6 $\pm$ 0.0\\
                                 & $\text{DeCoRe}_{\text{entropy}}$ & 68.2 $\pm$ 0.1 & 74.1 $\pm$ 0.1        & 65.1 $\pm$ 0.6\\
    \midrule
    \multicolumn{5}{c}{\textit{Llama3-8b-Instruct}} \\
    \midrule
    \multirow{4}{*}{ReAct}       & Standard                         & 62.2 $\pm$ 0.0 & 85.6 $\pm$ 0.0        & 36.4 $\pm$ 0.3\\
                                 & CAD                              & 50.6 $\pm$ 0.0 & 81.0 $\pm$ 0.3        & \textbf{44.4 $\pm$ 0.3}\\
                                 & DoLa                             & \textbf{62.6 $\pm$ 0.0} & 85.6 $\pm$ 0.0        & 36.1 $\pm$ 1.2\\
                                 & $\text{DeCoRe}_{\text{entropy}}$ & 54.4 $\pm$ 0.0 & \textbf{85.7} $\pm$ 0.2        & 39.3 $\pm$ 0.8\\ \bottomrule
    \end{tabular}
    \caption{Format Adherence (\%) grouped by model, framework, decoder and dataset. The highest adherence per group is in bold.}
    \label{tab:adherence}
\end{table}

\section{Prompts}
\label{App-Prompts}

\subsection{Direct}

\textit{Solve a question answering task. Answer in the format of: "the answer is: ANSWER." Here are some examples.}

\textit{Question: Who is the founder of the company which published Big Picture (Magazine)?}

\textit{The answer is: Sir Henry Wellcome.}

\textit{Question: Are Antoine Jean-Baptiste Thomas and Canardo (Rapper) of the same nationality?}

\textit{The answer is: yes.}

\textit{Question: Who is Elizabeth Stuart, Countess Of Lennox's mother-in-law?}

\textit{The answer is: Margaret Douglas.}

\textit{Question: When was Ludwig Gruno Of Hesse-Homburg's father born?}

\textit{The answer is: 19 May 1673.}

\textit{Question: Who was born first, Heo Keon or Ljiljana Nanušević?}

\textit{The answer is: Ljiljana Nanušević.}

\textit{Question: Which film was released more recently, The Black Duke or Avalalpam Vaikippoyi?}

\textit{The answer is: Avalalpam Vaikippoyi.}

\textit{Question: Which film has the director who is older, King Of The Zombies or Mon Oncle Benjamin?}

\textit{The answer is: King Of The Zombies.}

\textit{Question: Which film has the director who was born first, Dead Man'S Evidence or Creature Of The Walking Dead?}

\textit{The answer is: Dead Man's Evidence.}

(Llama only) \textit{Examples finished. Follow the exact format like shown in the examples.}

\subsection{OneR}

\textit{Solve a question answering task. You may get some context to help you answer the question if available. Answer in the format of: "the answer is: ANSWER." Here are some examples.}

\textit{Question: Who is the founder of the company which published Big Picture (Magazine)?}

\textit{Context: The Wellcome Trust is a research charity based in London, United Kingdom. It was established in 1936 with legacies from the pharmaceutical magnate Sir Henry Wellcome to fund research to improve human and animal health. The aim of the Trust is to "achieve extraordinary improvements in health by supporting the brightest minds", and in addition to funding biomedical research it supports the public understanding of science. It had a financial endowment of £25.9 billion in 2018, making it the fourth wealthiest charitable foundation in the world, after the American Bill \& Melinda Gates Foundation, the Danish Novo Nordisk Foundation and the Dutch INGKA Foundation (related to the IKEA company). The Trust has been described by the "Financial Times" as the United Kingdom's largest provider of non-governmental funding for scientific research, and one of the largest providers in the world.}

\textit{The answer is: Sir Henry Wellcome.}

\textit{Question: Are Antoine Jean-Baptiste Thomas and Canardo (Rapper) of the same nationality?}

\textit{Context: Hakim Mouhid( born 22 September 1984 in Trappes, Yvelines), better known by his stage name Canardo, is a French rapper, singer, songwriter and music producer. After working with label" Banlieue Sale Music" since 2007, in 2011, he founded his own label" Henijai Music". He is the brother of fellow French rapper La Fouine.}

\textit{The answer is: yes.}

\textit{Question: Who is Elizabeth Stuart, Countess Of Lennox's mother-in-law?}

\textit{Context: Charles Stuart, 1st (or 5th) Earl of Lennox (April or May 1557 – April 1576), was the fourth son of Matthew Stewart, 4th Earl of Lennox and Margaret Douglas, daughter of Margaret Tudor and granddaughter of King Henry VII of England. His brother was Henry Stuart, Lord Darnley.}

\textit{The answer is: Margaret Douglas.}

\textit{Question: When was Ludwig Gruno Of Hesse-Homburg's father born?}

\textit{Context: Frederick III James of Hesse-Homburg (born 19 May 1673 in Cölln; died: 8 June 1746 in 's-Hertogenbosch) was a Landgrave of Hesse-Homburg}

\textit{The answer is: 19 May 1673.}

\textit{Question: Who was born first, Heo Keon or Ljiljana Nanušević?}

\textit{Context: Ljiljana Nanušević( born 2 August 1981) is a Serbian former professional tennis player who played in the Fed Cup for Yugoslavia. Nanušević made all of her Fed Cup appearances in the 1997 Fed Cup, while aged only 15. Playing in a total of five ties, she played doubles in each of them, as well as singles once, managing to win every rubber. She won an ITF singles tournament in Istanbul in 2001 and won a further two ITF titles in doubles.}

\textit{The answer is: Ljiljana Nanušević.}

\textit{Question: Which film was released more recently, The Black Duke or Avalalpam Vaikippoyi?}

\textit{Context: Avalalpam Vaikippoyi is a 1971 Indian Malayalam film, directed by John Sankaramangalam and produced by United Producers. The film stars Prem Nazir, Sheela, Jayabharathi and Adoor Bhasi in the lead roles. The film had musical score by G. Devarajan.}

\textit{The answer is: Avalalpam Vaikippoyi.}

\textit{Question: Which film has the director who is older, King Of The Zombies or Mon Oncle Benjamin?}

\textit{Context: Mon oncle Benjamin ("My Uncle Benjamin") is a 1969 French film directed by Édouard Molinaro, starring Jacques Brel and Claude Jade. The film is based on a once-popular French comic novel "Mon oncle Benjamin" by (1842). The 1969 film "Don't Grieve", directed by the Georgian Georgi Daneliya, is also based on Tillier's novel as was Francis Bousquet's 1942 comic opera "Mon oncle Benjamin". The film was released on 28 November 1969.}

\textit{The answer is: King Of The Zombies.}

\textit{Question: Which film has the director who was born first, Dead Man'S Evidence or Creature Of The Walking Dead?}

\textit{Context: Creature of the Walking Dead is a 1965 horror film adapted by Jerry Warren from a 1961 Mexican horror film "La Marca del Muerto", which translates as "Mark of the Dead Man". The original Mexican film was directed by Fernando Cortés, written by Alfredo Varela Jr., and released in Mexico on October 12, 1961. The special effects was handled by Nicholas Reye.}

\textit{The answer is: Dead Man's Evidence.}

(Llama only) \textit{Examples finished. Follow the exact format like shown in the examples. The following question is the one you need to answer:}

\subsection{ReAct}

(2WikiMultihopQA and MuSique) \textit{Solve a question answering task with interleaving Thought, Action, Observation steps. Thought can reason about the current situation, and Action can be two types: (1) Search[entity], which searches Wikipedia and returns related paragraphs if they exist. (2) Finish[answer], which returns the answer and finishes the task. Here are some examples.}

(HotpotQA) \textit{Solve a question answering task with interleaving Thought, Action, Observation steps. Thought can reason about the current situation, and Action can be three types: (1) Search[entity], which searches the exact entity on Wikipedia and returns the first paragraph if it exists. If not, it will return some similar entities to search. (2) Lookup[keyword], which returns the next sentence containing keyword in the current passage. (3) Finish[answer], which returns the answer and finishes the task. Here are some examples.}

[8 Sets of examples, only one shown for brevity. Full examples in code provided.]

\textit{Question: Who is the founder of the company which published Big Picture (Magazine)?}

\textit{Thought 1: I need to search Big Picture (Magazine) to find out who founded the company that published it.}

\textit{Action 1: Search[Big Picture (Magazine)]}

\textit{Observation 1: The Big Picture series provides teachers and post-16 students with up-to-date information on research findings in biology and medicine, and the social and ethical implications of this research. Published by the Wellcome Trust as a free educational resource, each issue is available for free electronically. The website provides free resources for teachers and online activities for students, including lesson ideas, animations, image galleries and short videos. , the print subscription has been discontinued and replaced with online only.}

\textit{Thought 2: The Big Picture (Magazine) is published by the Wellcome Trust. I need to search the Wellcome Trust to find out who founded it.}

\textit{Action 2: Search[Wellcome Trust]}

\textit{Observation 2: The Wellcome Trust is a research charity based in London, United Kingdom. It was established in 1936 with legacies from the pharmaceutical magnate Sir Henry Wellcome to fund research to improve human and animal health. The aim of the Trust is to "achieve extraordinary improvements in health by supporting the brightest minds", and in addition to funding biomedical research it supports the public understanding of science. It had a financial endowment of £25.9 billion in 2018, making it the fourth wealthiest charitable foundation in the world, after the American Bill \& Melinda Gates Foundation, the Danish Novo Nordisk Foundation and the Dutch INGKA Foundation (related to the IKEA company). The Trust has been described by the "Financial Times" as the United Kingdom's largest provider of non-governmental funding for scientific research, and one of the largest providers in the world.}

\textit{Thought 3: The Wellcome Trust was established by Sir Henry Wellcome in 1936. Therefore, the founder of the company which published Big Picture (Magazine) is Sir Henry Wellcome.}

\textit{Action 3: Finish[Sir Henry Wellcome]}

(Llama only) \textit{Examples finished. Your thoughts should reason about the observations. Follow the exact format like shown in the examples. The following question is the one you need to answer:}

\end{document}